\documentclass[conference]{IEEEtran}

\usepackage[T1]{fontenc}
\usepackage{algorithm}
\usepackage{algorithmic}
\usepackage{blindtext, graphicx}
\usepackage{balance}
\usepackage{amsmath}
\usepackage{cite}
\usepackage[short]{optidef}
\usepackage{amssymb}
\makeatletter
\newcommand*{\rom}[1]{\expandafter\@slowromancap\romannumeral #1@}

\makeatother

\usepackage[]{caption}

\begin{document}

\title{Generative Adversarial Classification Network with Application to Network Traffic Classification \vspace{-3mm}}

\author{\IEEEauthorblockN{ Rozhina Ghanavi\IEEEauthorrefmark{1},
    Ben Liang\IEEEauthorrefmark{1},
    Ali Tizghadam\IEEEauthorrefmark{2},}
   
  \IEEEauthorblockA{\IEEEauthorrefmark{1}Dept. of Electrical and Computer Engineering,
    University of Toronto, Canada}
  \IEEEauthorblockA{\IEEEauthorrefmark{2}Technology Strategy and Business Transformation, TELUS Communications, Canada\\
  Email: rozhina.ghanavi@mail.utoronto.ca, liang@ece.utoronto.ca, ali.tizghadam@telus.com}
  \vspace{-10mm}}

\maketitle

\begin{abstract}\\
Large datasets in machine learning often contain missing data, which necessitates the imputation of missing data values. In this work, we are motivated by network traffic classification, where traditional data imputation methods do not perform well. We recognize that no existing method directly accounts for classification accuracy during data imputation. Therefore, we propose a joint data imputation and data classification method, termed generative adversarial classification network (GACN), whose architecture contains a generator network, a discriminator network, and a classification network, which are iteratively optimized toward the ultimate objective of classification accuracy. For the scenario where some data samples are unlabeled, we further propose an extension termed semi-supervised GACN (SS-GACN), which is able to use the partially labeled data to improve classification accuracy. We conduct experiments with real-world network traffic data traces, which demonstrate that GACN and SS-GACN can more accurately impute data features that are more important for classification, and they outperform existing methods in terms of classification accuracy.

\end{abstract}

\IEEEpeerreviewmaketitle

\section{Introduction}
Prediction, estimation, and inference using machine-learning approaches often rely on large and informative datasets. However, missing data is inevitable in many applications, due to faulty data collection, costly measurement, corrupted data storage, and other reasons \cite{BMar}. Some examples of fields where instances of missing data often occur are economics \cite{econ}, computational biology \cite{dna}, and medicine \cite{psycho}. Many published works in the field of machine learning and statistics cover missing data problems, such as \cite{Nash, Marlin1}. In \cite{bib}, the authors brought to light the problem of missing data in network traffic classification. They concluded that this issue is common in network traffic flow datasets. In network traffic classification, since the missingness often is spread widely throughout the datasets, the naive method of deleting all the data entries with at least one feature missing is not an option since that would delete a considerable portion of the dataset. Thus, we need to impute values in place of the missing data.
\let\thefootnote\relax\footnote{This work was funded in part by Telus Communications and the
Natural Sciences and Engineering Research Council (NSERC) of Canada.}

There is a large body of research on the imputation of missing data in machine learning. Classical methods include statistical solutions such as mean imputation and interpolation \cite{BMar}. However, these methods have been shown to be insufficient in modern applications. Subsequently, deep learning methods have been proposed. 
In \cite{bio2} the authors focused on finding a discriminative approach that can handle missing features in all types of data, continuous or categorical, simultaneously. To this end, they proposed an iterative method based on a random forest model.
In \cite{mice1, mice2}, an imputation method based on the chained equation was presented, termed multivariate imputation by chained equations (MICE). It is a multiple imputation approach, which uses the best imputation solution chosen from a candidate set, for individual columns features. MICE is one of the state-of-the-art solutions and has a popular software implementation that is widely used in the literature.

Deep generative imputation methods have attracted much attention in recent years \cite{MissVAE, dataWig, missnew}. The main benefit of using generative models is that they make the uncertainty estimation of imputed value possible with multiple imputation \cite{multi}. In generative adversarial imputation networks (GAIN) \cite{GAIN}, the authors proposed a model based on the generative adversarial networks (GAN) \cite{GANs} for imputation of missing data entries in tabular datasets. The GAN architecture consists of a generator and a discriminator. The generator is a multilayer perceptron generating fake samples, and the discriminator is another multilayer perceptron trying to maximize the probability of assigning the right label to the observed (real) data or generated (fake) data. GAN aims to teach the generator to produce real looking samples by modeling the optimization problem as a two-player minimax game between the generator and the discriminator. Borrowing GAN's idea, GAIN imputes missing data by generating the missing parts in data entries, where the problem is formulated as a two-player game between the generator and discriminator. The generator generates the data entries. The discriminator checks every single entry for whether it is imputed or observed. In \cite{HEXAGAN} the authors further proposed a data pre-possessing method based on \cite{GAIN} for missing data imputation and handling imbalanced datasets.

None of the existing methods directly account for data classification during data imputation. As a typical example, in the research work on GAIN, the missing feature data were imputed first, with the average root mean squared error (RMSE) as a main performance metric. The accuracy of classification was considered only in experimental performance evaluation, which was separate from data imputation. Therefore, in applications where data classification is the ultimate objective, e.g., network traffic identification \cite{bib}, existing data imputation methods suffer a loss of efficacy by not directly aligning their objectives with classification accuracy. 

In this work, we address the above challenge by jointly considering data imputation and data classification. A unique characteristic of our work is that we take into account the classification accuracy as our primary motivation in data imputation. Our main contributions are as follows:

\begin{itemize}

\item We propose a new generative model for imputing missing data features, termed generative adversarial classification network (GACN), which consists of three inter-connected deep neural networks, a generator network, a discriminator network, and a classification network. We design a weighted loss function based on the three networks and an iterative three-step optimization algorithm to train GACN toward improving the classification accuracy.

\item We further propose an extension, termed semi-supervised GACN (SS-GACN), which does not require all data samples to be labeled as GACN does. Therefore, SS-GACN allows missing values even in the data labels. It may be viewed as a general form of both GACN and GAIN.

\item As an application to network traffic classification, we perform experiments with real-world data traces using a combination of ISCX VPN-nonVPN \cite{iscx1, iscx2} and ISCX Tor-nonTor \cite{iscx3, iscx4} datasets. Comparison with GAIN, MICE, and mean imputation demonstrates that GACN and SS-GACN can more accurately impute data that are more important for classification. They achieve higher classification accuracy faster, for a wide range of experimental settings.

\end{itemize}

The rest of this paper is organized as follows. Section~\ref{section:classifier} presents our system and problem statement, including an application of network traffic application. In Section~\ref{section:framework}, we present the GACN architecture and optimization algorithm for joint data imputation and data classification. The approach we propose aims for better classification results after imputation is completed. Section~\ref{section:semisup} presents the SS-GACN extension of our proposed method, which enables us to work with a partially labeled dataset. Section~\ref{section:Data} is devoted to presenting the experimental results. Finally, Section~\ref{section:conclusion} concludes this paper.

\vspace*{-1.5mm}
\section{System Model and Problem Statement}
\vspace*{-1.5mm}
\label{section:classifier}
Our goal is to maximize classification accuracy, given any real-world dataset which consists of missing data feature values. To this end, we define our problem as follows. 

Suppose data vector, $X = (X_1, X_2, \ldots, X_d)$, is a random vector in $\mathbb{R}^{d}$. Every $X$ is labeled with a target value, $T$, indicating the class to which it belongs. Let $M = (M_1, M_2, \ldots, M_d)$ in $\{0,1\}^d$ be a random vector, which we call the \textit{missingness vector}, so that if a feature value $X_i$ is missing, $M_i=0$.
Our goal is to impute missing entries of ${X}$, with an imputation algorithm in a way that maximizes the \textit{classification accuracy}. Note that the imputation algorithm should be such that it outputs and imputes value for $X_i$ if and only if $M_i=0$; otherwise, we have already observed $X_i$ and its value should be retained.

We consider a training dataset with $N$ samples,  $\mathcal{D} = \{ (x^1, t^1), (x^2, t^2), \ldots, (x^N, t^N) \}$, where each sample is drawn independently from some distribution of $X$ with arbitrary missing elements.\footnote{In general, we use uppercase letters to represent random variables and lowercase letters to represent the realization of random variables.} For the classification accuracy metric, we consider the cross-entropy loss.
It is noteworthy that any general metric can be used instead of cross-entropy. However, if another metric is chosen, then the proposed algorithm in Section~\ref{section:framework} needs to be modified accordingly.

As an application for the proposed methods, in Section~\ref{section:Data}, we consider a network traffic flow dataset that combines ISCX VPN-nonVPN \cite{iscx1, iscx2} and ISCX Tor-nonTor \cite{iscx3, iscx4} datasets. The objective is to identify whether a flow is delay-sensitive or delay-tolerant. In other words, the goal is to classify traffic flows based on their quality-of-service (QoS) under the presence of missing data.

\section{GACN Framework}
\label{section:framework}
With missing data features, optimal classification implies the need for imputing those missing data features. Instead of the conventional approach of separately imputing data and then performing classification, we propose the generative adversarial classification network (GACN) for the imputation of missing data features while considering the classification accuracy. The architecture of GACN consists of three neural networks: the generator network $G$, the discriminator network $D$, and the classification network $A$. We next present the details of GACN.

\subsection{The Generator Network}
The definition of the generator is as follows. We first define a noise vector, $Z \in [0, 1]^d$. This noise vector is the input to the generator. We express the generator as $G(X,M,Z;\theta_g)$, a differentiable function taking value in $\mathbb{R}^{d}$, which is a multilayer perceptron with parameters $\theta_g$. It takes ${M} \odot {X}+({1}-{M}) \odot {{Z}}$ as input, where $\odot$ denotes element-by-element multiplication. It outputs a vector of imputed feature values:
\begin{IEEEeqnarray}{c}
\check{X} = G(X, M, Z)
\IEEEeqnarraynumspace
\vspace{-2mm}
\label{eq:labelc1}
\end{IEEEeqnarray}

Notably, $G$ generates a value for all the feature entries, both observed and unobserved. However, if a value is observed, we use the observed value as the algorithm's output. Thus it is essential to update the output as
\begin{equation}
\hat{X}_i = \left\{ \,
\begin{IEEEeqnarraybox}[][c]{l?s}
\IEEEstrut
X_{i} & if $M_{i}=1$, \\
\check{X}_i & $\text{otherwise}$.
\IEEEstrut
\end{IEEEeqnarraybox}
\right.
\vspace{-2mm}
\label{eq:output_update}
\end{equation}
Now the entries of the overall output vector $\hat{X}$ are equal to those of $X$ for the observed values and are equal to the generated values for missing features.

\subsection{The Discriminator Network}
\label{section:discriminator}
The discriminator, $D$, is one of the mechanisms to check whether the output of the generator $G$ is similar to previously learned data pattern. It is a multilayer perceptron denoted by $D(\hat{X}; \theta_d)$, with the network parameters $\theta_d$. It outputs a vector in the $d$-dimensional region $[0, 1]^d$, which represents the probabilities of the data entries in $\hat{X}$ being observed instead of imputed. Inspired by \cite{GAIN}, for correct operation, it is necessary to define a selection vector, $R \in \{0,1\}^d$. This is because learning a good discriminator in our problem is harder than in basic GAN. Here, the discriminator needs to assign the probability of being imputed to every single feature, while GAN's discriminator only needs to decide if the entire generated set is real or fake. The selection vectors gives the training of $D$ some information on the likelihood that a feature value is missing for any given data sample, to make this job easier. The selection vector $R$ is define as follows:

\begin{equation} R_{i}=\left\{\begin{array}{l}
1 \text { if } i \neq r \\
0 \text { if } i=r,
\end{array}\right.\vspace{-1mm}\end{equation}
where $r$ is uniformly sampled from ${1, \ldots, d}$.
With $R$, now the discriminator needs to decide only whether the feature  $i=r$ is imputed or observed. Furthermore, for each data sample $X$, since we know which data features are missing, we update the output of the discriminator as follows:
\vspace{-2mm}

\begin{equation}
\hat{P}(X, R) = \left\{ \,
\begin{IEEEeqnarraybox}[][c]{l?s}
\IEEEstrut
D(X)_{i} & if $R_{i}=0$, \\
M_i & $\text{otherwise}$.
\IEEEstrut
\end{IEEEeqnarraybox}
\right.
\label{eq:p_hat}
\end{equation}

\vspace{-1.5mm}
\subsection{The Classification Network}
The classification network, $A(\hat{X};\theta_a)$, is the last multilayer perceptron of the GACN model with parameters $\theta_a$. It is a conventional classification network that outputs the probability of assigning the data vector $\hat{X}$ to a specific target label.

\vspace{-1mm}
\subsection{Connecting the Components of GACN}
Fig. 1 shows how these three networks interact with each other. As shown in this figure, $G$ receives five elements as its input: the data vector $X$, the noise vector $Z$, the missingness vector $M$, and the outputs of $D$ and $A$. The generator then outputs the completed vector $\hat{X}$. The completed vector is then one of the inputs to the discriminator. In addition to the completed data vector, $D$ also receives the selection vector and outputs the probability vector of each data entry being imputed. The completed vector from the generator is also the input to the classification network, $A$. The classification network outputs the probability of assigning each data points to different classes.

\begin{figure}[t]
  \includegraphics[width=0.35\textwidth]{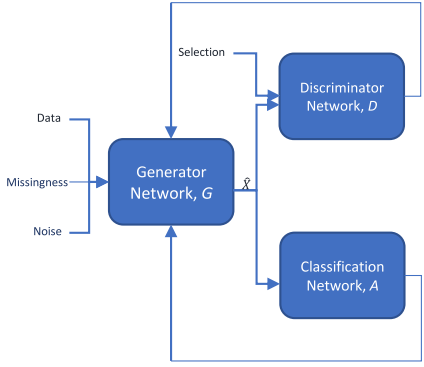}
  \centering
  \caption{GACN model schematic.}
  \vspace{-5mm}
  \label{GACN_sch}
\end{figure}

As in GAN, we form a game to train a generator that outputs artificial values for missing data that both match the learned data pattern and provide high classification accuracy. Thus, in GACN, the generator $G$ generates replacements for the missing data. Then the discriminator $D$ gives feedback about how good the generator's imputation is, but only in terms of the previously learned data pattern. Finally, the classification network $A$ checks how much the imputation helps in terms of the final accuracy. The game between the generator and the discriminator occurs sequentially in iterations over time. The classification network is not a player, but it actively gives feedback to help the generator and the discriminator find which feature is more critical in terms of accuracy. This feedback makes sure that for more essential features, the data imputation is performed more carefully. 

Similar to basic GAN, for the sequential game between the generator and the discriminator, the aim is to solve the following minimax problem.
\begin{equation}\min _{\theta_g} \max _{\theta_d} L(D, G),
\vspace{-1mm}
\label{eq:min_max}
\end{equation}
where $L(D, G)$ is defined as follows:

\begin{equation}\begin{aligned}
L(D, G)=\mathbb{E}_{ X, M, R}[M^T \log \hat{P}(\hat{X}, R) + \\
(1-M)^T \log (1-\hat{P}(\hat{X}, R))].
\vspace{-3mm} \end{aligned}
\label{eq:LDG}\end{equation}
In \eqref{eq:LDG}, the discriminator network tries to maximize the probability of correctly deciding whether a feature is real or imputed. The generator, on the other hand, tries to minimize the chance of the discriminator deciding correctly. This implicitly means that the generator's goal is to impute the data so well that the imputed features are indistinguishable from the real ones. Here, $L(D, G)$ is an expectation over three random variables' realization. The first two random variables, $X$ and $M$, are the data vector and missingness vector. These two random variables are defined in Section~\ref{section:classifier}. The last random variable $R$ is the selection vector presented in Section~\ref{section:discriminator}. As mentioned earlier, this random variable's existence is essential for assuring adequate performance from the discriminator. The dependence of \eqref{eq:LDG} to the generator is trough $\hat{X}$. The first part of this equation, $M^T \log \hat{P}(\hat{X}, R)$, checks how well the discriminator assigns the observed values' probabilities. At the same time, the second part checks the same for missing and imputed values.

\vspace*{-0.5mm}
\subsection{GACN Algorithm}
\vspace*{-0.5mm}
We next discuss how to address the minimax problem in \eqref{eq:min_max} while accounting for classification accuracy. First, we can extend the analysis of \cite{GAIN} to show that there exists a global optimum for \eqref{eq:min_max}, and this global optimum is the true data distribution. The formal proof is omitted for brevity.

Our method is based on three-step optimization. In the following, $\alpha, \beta$, and $\gamma$ are model hyperparameters, and $B_D, B_A,$ and $B_G$ are mini-batch sizes. Also, superscript $j$ denotes the $j$-th sample in a mini-batch. We first update $D$ and $A$ given a fixed generator $G$. For updating $D$, we first define the cross-entropy for a sample being observed as follows:
\begin{equation}\begin{aligned}
    \mathcal{L}_{D}(m, \hat{P}(\hat{x}, h)) = 
    \sum_{i=1}^{d}m_{i} \log (\hat{P}(\hat{x}, h)_i)\\
+(1-m_{i}) \log (1-\hat{P}(\hat{x}, h)_i).\vspace{-1.5mm}
\end{aligned}\end{equation}
Then $D$ is trained with the following objective:
\begin{equation}
    \min _{\theta_{d}} \quad -\sum_{j=1}^{B_{D}} \mathcal{L}_{D}(m^j, \hat{P}(\hat{x}^j, h^j)).
\vspace{-1.2mm} \end{equation}
For updating $A$ we define $\mathcal{L}_{A}(\hat{x}^j)$ as the cross-entropy loss between the target value $t$ and the output probability vector $A(\hat{x})$. Then we optimize $A$ with respect to the following objective:
\begin{equation}\begin{aligned}
    \min_{\theta_{a}} \quad -\sum_{j=1}^{B_{A}} \mathcal{L}_{A}(\hat{x}^j).
\end{aligned}\vspace{-1.5mm}\label{eq:La}\end{equation}
Now with the given locally optimal $A$ and $D$, we can update $G$. We use a weighted loss function consisting of three elements, $\mathcal{L}_{M}$, $\mathcal{L}_{G}$, and $\mathcal{L}_{P}$, which are defined as follows:    
\begin{equation}\mathcal{L}_{M}(x, \hat{x})=\sum_{i=1}^{d} m_{i}(x_{i} - {\hat{x}_{i}})^2\vspace{-2mm}\end{equation}
\begin{equation}\begin{aligned}
\mathcal{L}_{G}(m, \hat{P}(\hat{x}, {h}))=
-\sum_{i=1}^{d}(1-m_{i}) \log (\hat{P}(\hat{x}, h)_i),
\end{aligned}\vspace{-2mm}\end{equation}
\begin{equation}\mathcal{L}_{P}(m, \hat{x})=- \log(A(\hat{x}))\sum_{i=1}^{d}(1-m_{i}).\vspace{-1.5mm}
\end{equation}
Minimizing $\mathcal{L}_{M}$ makes sure the generator learns to generate values close to the observed data's real values. While optimizing $\mathcal{L}_{G}$ and $\mathcal{L}_{P}$ helps better impute missing values. $\mathcal{L}_{G}$ is the loss function associated with fooling the discriminator. Optimizing this loss function means the generator is so good at imputing missing values that the discriminator cannot distinguish between imputed or real features. The last loss function, $\mathcal{L}_{P}$, is a term associated with the classification loss. Adding this term helps the generator learn the model, which gives the highest accuracy by emphasizing the features that are more important in terms of final accuracy.

Given these three loss functions we then update the generator based on the following objective:
\begin{equation}\begin{aligned}
\min_{\theta_{g}} \sum_{j=1}^{B_{G}} \alpha \mathcal{L}_{M}(x^j, \hat{x}^j) + \beta \mathcal{L}_{G}(m^j, \hat{P}(\hat{x}^j, h^j)) \\
+
\gamma  \mathcal{L}_{P}(m^j, \hat{x}^j).\vspace{-3mm}
\end{aligned}\label{eq:Ltot}\end{equation}
Algorithm 1 presents the pseudo-code of our algorithm.

\vspace{-0.75mm}
\section{Semi-Supervised GACN}
\vspace{-0.75mm}
\label{section:semisup}
GACN requires that all data samples are labeled to run correctly. However, as shown in Section~\ref{section:Data}, if we do not have enough labeled samples, the GACN performance drops drastically. Therefore, an extension of GACN is presented here to address this issue. We call this algorithm semi-supervised GACN (SS-GACN). Alternatively, we may view SS-GACN as a more robust version of GACN, which allows missing data labels in addition to missing data features.

In SS-GACN, we update \eqref{eq:La} and \eqref{eq:Ltot} as follows:
\begin{equation}\begin{aligned}
    \min_{\theta_{a}} -\sum_{j=1}^{B_{A}} \kappa^j \mathcal{L}_{A}(\hat{x}^j),
\vspace{-3mm}\end{aligned}\end{equation}
\begin{equation}\begin{aligned}
\min_{\theta_{g}} \sum_{j=1}^{B_{G}} \alpha \mathcal{L}_{M}(x^j, \hat{x}^j) + \beta \mathcal{L}_{G}(m^j, \hat{P}(\hat{x}^j, h^j)) \\
+\kappa^j \gamma  \mathcal{L}_{P}(m^j, \hat{x}^j),
\vspace{-3mm}\end{aligned}\label{Ltot1}\end{equation}
where $\kappa^j$ is a binary variable that is equal to 1 if the label for the $j$-th sample in the mini-batch is present and 0 otherwise. For labeled samples, SS-GACN uses the additional information in the labels to update the classification network $A$. However, if a sample is unlabeled, it still helps to update the discriminator and also the generator by updating $\mathcal{L}_{D}$, $\mathcal{L}_{M}$, and $\mathcal{L}_{G}$ using this sample. We note that SS-GACN is a general algorithm, of which both GACN and GAIN are special cases, when all data samples are labeled and when there is no labeled sample, respectively. In Section~\ref{section:Data}, we show that in the presence of partially labeled data, SS-GACN outperforms both GACN and GAIN.

\setlength{\textfloatsep}{1pt}
\begin{algorithm}[htbp]
\begin{algorithmic}
\FOR{a preset number of iterations}
\STATE \textbf{(1) Discriminator optimization}
\STATE Sample from the dataset $\{(x^j, m^j)\}_{j=1}^{B_{D}}$ 
\STATE Sample i.i.d., $\{z^j\}_{j=1}^{B_{D}}, \text { of } Z$
\STATE Sample i.i.d., $\{r^j\}_{j=1}^{B_{D}}, \text { of } R$
\FOR {$j=1, \ldots, B_{D}$}
\STATE $\check{x}^j \leftarrow G(x^j, m^j, z^j)$ 
\FOR{$i$ in $d$}
\IF{$M_{i}=1$}
\STATE $\hat{x}^j _i = x^j_i$
\ELSE
\STATE $\hat{x}^j _i = \check{x}^j_i$
\ENDIF
\ENDFOR
 \STATE $h^j=r^j \odot m^j+0.5(1-r^j)$
\ENDFOR
\STATE Optimize $D$ with respect to objective
\STATE $
\nabla_{\theta_d}-\sum_{j=1}^{B_{D}} \mathcal{L}_{D}(m^j, \hat{P}(\hat{x}^j, h^j)) 
$

\STATE \textbf{(2) Classification network optimization}
\STATE Sample from the dataset $\{(x^j,  m^j)\}_{j=1}^{B_{A}}$
\STATE Sample i.i.d., $\{ z^j\}_{j=1}^{B_{A}}, \text { of }  Z$
\FOR{$j=1, \ldots, B_{A}$}
\STATE $\check{  x}^j \leftarrow G(\tilde{ x}^j,  m^j,  z^j)$
\FOR{$i$ in $d$}
\IF{$M_{i}=1$}
\STATE $\hat{x}^j _i = x^j_i$
\ELSE
\STATE $\hat{x}^j _i = \check{x}^j_i$
\ENDIF
\ENDFOR
\ENDFOR
\STATE Optimize $A$ with respect to objective
\STATE $\nabla_{\theta_a} -\sum_{j=1}^{B_{A}} \mathcal{L}_{A}(\hat{x}^j)$

\STATE \textbf{(3) Generator optimization} \\
\STATE Sample from the dataset $\{(\tilde{ x}^j,  m^j)\}_{j=1}^{B_{G}}$
\STATE Sample i.i.d., $\{ z^j\}_{j=1}^{B_{G}}, \text { of }  Z$
\STATE Sample i.i.d., $\{ b^j\}_{j=1}^{B_{G}}, \text { of   R}$
\STATE Optimize $G$ with respect to objective
\STATE $
\nabla_{\theta_g} \sum_{j=1}^{B_{G}} \alpha \mathcal{L}_{M}(x^j, \hat{x}^j) + \beta \mathcal{L}_{G}(m^j, \hat{P}(\hat{x}^j, h^j)) + \gamma  \mathcal{L}_{P}(m^j, \hat{x}^j)
$

\caption{Pseudo-code of GACN }
\ENDFOR
\end{algorithmic}
\end{algorithm}

\section{Experiments in Network Traffic Classification}
\label{section:Data}
To experiment with the proposed GACN algorithm in the application of network traffic classification, we consider a combination of ISCX VPN-nonVPN \cite{iscx1, iscx2} and ISCX Tor-nonTor \cite{iscx3, iscx4} datasets. These datasets are PCAP files from encrypted TCP flows, each labeled with one of 22 applications. Following the procedure in \cite{sayantan}, we extract 266 features for each flow data sample from the dataset. We consider binary QoS classification of the flows. To create labels, we map each of the applications to the delay-sensitive group or the delay-tolerant group (e.g., video conference is delay sensitive, email is delay tolerant, etc.). This forms the \textit{flow dataset}. It consists of 43590 flows.

In order to experiment with different levels of missing data, we first take a complete dataset, i.e., with no missing data, based on the flow dataset and then add artificial random missingness on its data. To build the complete dataset, we delete all the samples with at least one unobserved feature. Then, we upsample the smaller classes to keep the classification fair between elements. The final dataset contains 34446 flows. Now with this dataset, we use 70\% of the data for training, and 10\% and 20\% for the validation set and test set, respectively. We assume that each feature $X_i$ is independently missing with probability $P_{\textrm{miss}}$, which corresponds to the missing at random (MAR) model \cite{nmar}.

In our experiments $G$ and $D$ are three-layer perceptron networks, where all hidden layers have nodes equal to the numbers of features, while $A$ is another three-layer perceptron, where the hidden layers consist of 30, 20, and 20 nodes respectively. All networks are optimized with ADAM \cite{Adam}. We choose these architectures based on extensive experimentation and hyper-parameter tuning. 

For performance comparison, we consider GAIN \cite{GAIN}, MICE \cite{mice1, mice2}, and mean imputation, a popular method in practical systems where for each feature, the average of all observed values of the feature in the dataset is used as the value of the same feature in samples where the feature value is missing. Since the samples are already balanced, we use the test accuracy for performance metric. We note that throughout this section, all numerical results include 90\% confidence intervals for 50 random realizations.

\vspace*{-2mm}
\subsection{Improved Classification Accuracy}
\begin{figure}[t]
  \includegraphics[width=0.4\textwidth]{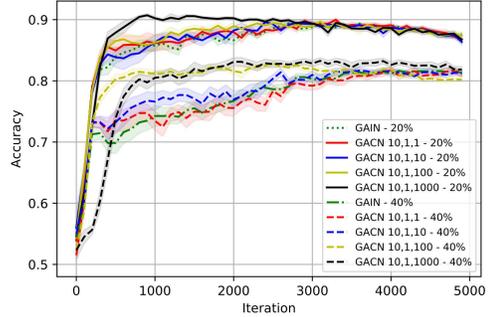}
  \centering
  \caption{Comparison of test accuracy between GACN and GAIN. The four numbers in the legend are the values of $\alpha, \beta, \gamma,$ and $P_{\text{miss}}$.}
  \label{accu20}
\end{figure}

\vspace*{-1mm}
Fig. \ref{accu20} studies the effectiveness of having the classification loss $\mathcal{L}_P$ as a part of the generator's update with $P_{\textrm{miss}}$ equal to 20\% and 40\%. These missingness rates are common choices in similar studies (e.g. \cite{GAIN, MissVAE}). We have used $\alpha=10$ and $\beta=1$ following the recommendation for GAIN in \cite{GAIN}. Fig. \ref{accu20} shows that by picking a suitable $\gamma$, GACN can reach a higher accuracy faster than GAIN. For the remaining results, we use $\gamma=1000$ as default, unless otherwise specified.

Table I gives further comparison between GACN and GAIN for different missingness rates. The number of iterations is 1000. We can see that GACN always gives higher classification accuracy with a tight confidence interval. This demonstrates the advantage of GACN by integrating data imputation and classification. Note that when $P_{\text{miss}}$ is low, both algorithms perform well as expected, and when $P_{\text{miss}}$ is above $50\%$, both algorithms suffer from the large amount of missing data. However, over a large range of moderate $P_{\text{miss}}$ values, GACN substantially outperforms GAIN.

Table II compares the accuracy results for different imputation methods, including GACN, GAIN, MICE, and mean imputation. The number of iterations for GACN and GAIN is 1000. We observe that GACN outperforms all other methods.

\begin{table}[btp]
\centering
\begin{tabular}{ |c|c|c| } 
 \hline
$P_{\text{miss}}$ & Accuracy $\textrm{GACN}$ & Accuracy \text{GAIN}  \\ 
 \hline
10\% & 0.9376 $\pm$ 0.0036 &  0.9237 $\pm$ 0.0052\\
\hline
20\% & 0.9022 $\pm$ 0.0024 & 0.8575 $\pm$ 0.0049\\
\hline
30\% & 0.8637 $\pm$ 0.0023 & 0.8326 $\pm$ 0.0046\\
\hline
40\% & 0.8017 $\pm$ 0.0028 & 0.7258 $\pm$ 0.0100\\
\hline
50\% & 0.7307 $\pm$ 0.0079 & 0.6886 $\pm$ 0.0072\\
\hline
60\% & 0.6389 $\pm$ 0.0100 & 0.6271 $\pm$ 0.0050\\
\hline
\end{tabular}
\caption{Test accuracy for GACN and GAIN at different missingness rates.} \label{tab:title} 
\end{table}

\vspace{-0.5mm}
\begin{table}[btp]
\centering
\begin{tabular}{ |c|c| } 
 \hline
Algorithm & Accuracy  \\
 \hline
 \text{GACN} & 0.9022 $\pm$ 0.0024\\
\hline
\text{GAIN} & 0.8575 $\pm$ 0.0049 \\
\hline
\text{MICE} & 0.8349 $\pm$ 0.0043\\
\hline
\text{Mean imputation} & 0.7540 $\pm$ 0.0016\\
\hline
\end{tabular}
\caption{Test accuracy for different algorithms at 20\% missingness rate.} \label{tab:title} 
\end{table}

\vspace{-1mm}
\subsection{Imputation RMSE}
\begin{figure}[t]
  \includegraphics[width=0.4\textwidth]{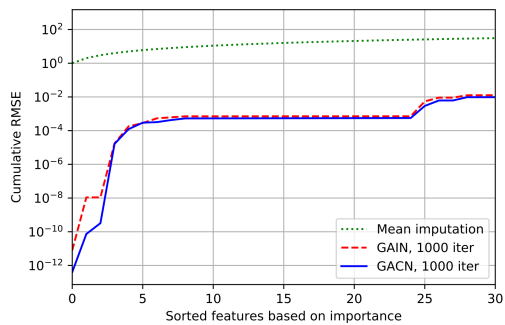}
  \centering
  \caption{Cumulative RMSE for the 30 most important features, sorted in decreasing importance.}
  \label{RMSE_fewIn}\vspace{-0.5mm}
\end{figure}

\vspace{-0.5mm}
To understand why GACN outperforms GAIN in terms of accuracy, we study the average RMSE of individual imputed features. We first sort all features based on importance. Here, feature importance is defined by the correlation between each feature and true labels in the complete dataset. The bigger the absolute value of the correlation coefficient, the higher the importance of the feature is. In Fig. \ref{RMSE_fewIn}, we plot the cumulative RMSE for the 30 most important features, for GACN, GAIN, and mean imputation. Each data point is the average of 30 realizations. We can see that GACN tends to impute more important features more accurately. This result shows that GACN puts in an additional effort by using the term $ \mathcal{L}_{P}$ to make sure the classification accuracy is high while imputing the missing data.

\vspace{-0.5mm}
\subsection{Partially Labelled Data Imputation}
\begin{figure}[t]
  \includegraphics[width=0.4\textwidth]{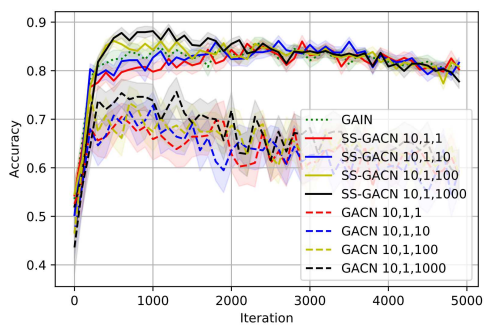}
  \centering
  \caption{SS-GACN and GACN test accuracy for $P_{\text{miss}} = 20\%$. The three numbers in the legend are the values of $\alpha, \beta, \gamma$.}
  \label{SSGACN_20}\vspace{+0.5mm}
\end{figure}

\vspace{-0.5mm}
We experiment with SS-GACN for partially labeled data. Fig. \ref{SSGACN_20} compares the accuracy result for GACN, SS-GACN, and GAIN when 20\% of the feature values are missing, and only 10\% of the training dataset is labeled. We can see that the performance of GACN drops drastically, because GACN can only use labeled data. On the other hand, GAIN preserves its performance because it does not use any labels. However, SS-GACN outperforms both of GACN and GAIN because it can improve its learning using the available 10\% labels and use the whole dataset for imputation.
\vspace{-0.5mm}
\vspace{-0.5mm}
\vspace{+0.5mm}
\section{Conclusion}
\label{section:conclusion}
We propose a novel deep generative method for the imputation of missing data features, termed GACN, that takes classification accuracy into account. The architecture of GACN consists of three neural networks working together to learn the data distribution, impute missing values, and perform classification. Our experimental result on real-world network traffic data traces show the performance advantage of GACN over the state of the art under a wide range of scenarios. We conclude that GACN achieves higher classification accuracy by more carefully imputing the data features that are more important for classification. We further expand our proposal to SS-GACN for partially labeled datasets. SS-GACN is able to use both unlabeled and labeled data entries, making it more useful in applications where it is hard to collect all the labels. Our experimental results further show that SS-GACN maintains satisfactory classification accuracy even when only a small percentage of data samples are labeled.

\balance

\end{document}